\documentclass{article}

\usepackage{arxiv}

\usepackage[utf8]{inputenc} 
\usepackage[T1]{fontenc}    
\usepackage{hyperref}       
\usepackage{url}            
\usepackage{booktabs}       
\usepackage{amsfonts}       
\usepackage{nicefrac}       
\usepackage{microtype}      
\usepackage{lipsum}
\usepackage{graphicx}
\usepackage{bm}
\usepackage{array}
\usepackage{mathtools}
\usepackage{bbm}
\usepackage{algorithm}
\usepackage{algpseudocode}
\usepackage{amsfonts}
\usepackage{booktabs}
\usepackage{makecell}
\usepackage{multirow}
\usepackage{subfigure}
\usepackage{ulem}
\usepackage{caption}
\usepackage{tabularx}
\usepackage[export]{adjustbox}

\graphicspath{ {./images/} }

\title{ LookHops: light multi-order convolution and pooling for graph classification }
\newcommand{\printfnsymbol}[1]{
  \textsuperscript{\@*}
 }
\author {
      Zhangyang Gao \thanks{These authors contribute equally.}  \textsuperscript{ ,} \textsuperscript{\rm 1},
        Haitao Lin\footnotemark[1]  \textsuperscript{ ,} \textsuperscript{\rm 1},
        Stan. Z Li \thanks{Corresponding author.} \textsuperscript{ ,}    \textsuperscript{\rm 1} \\

     \textsuperscript{\rm 1} Center for Artificial Intelligence Research and Innovation, Westlake University. \\
    $\{$gaozhangyang, linhaitao, stan.zq.li$\}$@westlake.edu.cn
}

\begin{document}
\maketitle
\begin{abstract}
Convolution and pooling are the key operations to learn hierarchical representation for graph classification, where more expressive $k$-order($k>1$) method requires more computation cost, limiting the further applications. In this paper, we investigate the strategy of selecting $k$ via neighborhood information gain and propose light $k$-order convolution and pooling requiring fewer parameters while improving the performance. Comprehensive and fair experiments through six graph classification benchmarks show: 1) the performance improvement is consistent to the $k$-order information gain. 2) the proposed convolution requires fewer parameters while providing competitive results. 3) the proposed pooling outperforms SOTA algorithms in terms of efficiency and  performance.

\end{abstract}

\section{Introduction}
Stacked convolution and pooling layers enable Convolutional Neural Networks (CNNs) to learn hierarchical representation of grid-like data\cite{krizhevsky2017imagenet}, where the convolution extracts local patterns of the data and the pooling layers reduce the computation cost by compressing the data shape. Because both of the two operations are defined on planar grids in Euclidean domains, they cannot be directly employed in graph data, which is a more general case and widely used in fields of chemical molecules, drug design and social networks. Learning the hierarchical representation of graph is a challenging problem and one of the solutions is to extend the convolution and pooling to graph.

Graph convolution includes spatial and spectral methods\cite{defferrard2016convolutional,atwood2016diffusion}, both of which can be seen as a message passing process on multi-hop graphs. For implementation on graphs of massive number of nodes, 1-order convolution, represented by GCN and GAT\cite{kipf2016semi,velivckovic2017graph}, become increasingly popular, but abandon part of ability to capturing complex graph pattern. For example, recent works have shown the limitation of 1-order convolution on the task of classifying non-isomorphism graphs \cite{morris2019weisfeiler}. Thus, efficient yet expressive $k$-order convolution should be considered for graph classification.



Graph pooling aims to solve the problem that GNNs are difficult to learn graph-level features in a hierachical way instead of simply extracting flat node-level features, as well as to reduce the computation costs on large graphs. Generally speaking, node dropping and structure learning are the two modules of  pooling, widely used in TopkPool\cite{gao2019graph},HGPSL\cite{zhang2019hierarchical}, SAGPool\cite{lee2019self}, iPool\cite{gao2019ipool}, ASAP\cite{ranjan2020asap} and EdgePool\cite{diehl2019edge}, where learnable scores are used to drop nodes(or edges). SAGPool and ASAP demonstrate that node scores learned from multi-order information improve the performance, but require too much computation. Besides, few works jointly consider structure learning together with pooling.

In this work, we explore the problem of designing efficient and effective $k$-order($k>1$) graph convolution and pooling. For reducing the computation, we also provide the strategy of selecting smaller $k$ with comparably extensive view of information. Evaluated results on six graph classification benchmarks show that the proposed methods can provide competitive results to the state-of-the-art algorithms, while requiring much fewer parameters. Our contributions include:
\begin{enumerate}
    \item Propose a criterion of selecting $k$ for $k$-order graph convolution and empirically prove the validity.
    \item Propose a light $k$-order convolution to provide competitive performance with fewer parameters.
    \item Propose a $k$-order pooling, outperforming the state-of-the-art algorithms both in efficiency and performance.
\end{enumerate}

\section{Related Work}
\paragraph{Graph convolution} Graph convolution performs message passing on the graph structure, which can be classified as spectral\cite{bruna2013spectral} and spatial\cite{atwood2016diffusion,niepert2016learning,gilmer2017neural} based methods. The SpectralCNN\cite{bruna2013spectral} requires expensive computation due to the eigen-decomposition of graph Laplacian. ChebyConv \cite{defferrard2016convolutional} solved the issue by using Chebyshev polynomials of Laplacian to approximate graph signals. Then, Kipf\cite{kipf2016semi} further simplified the $k$-order ChebyConv to 1-order GCN for scaling to large graph. Due to the limited ability to handling directed graph and generalizing to different structure, spatial GNNs are further recommended. For example, DCNN\cite{atwood2016diffusion} uses diffusion process modeling graph convolution, GraphSAGE\cite{hamilton2017inductive} learns the neighborhood aggregating function for extending to various graph structure and GAT\cite{velivckovic2017graph} can handle directed graph by aggregating neighborhood information through the attention mechanism.

\paragraph{Graph pooling} Graph pooling\cite{ying2018hierarchical} contains both node dropping and structure learning functions, aiming to extract hierarchical graph-level features and reduce the computation costs by compressing the size of the input data. Two types of scoring methods, based on edge or node, can be used for dropping nodes. EdgePool\cite{diehl2019edge} extracts the graph topology by merging endpoint of the high-scoring edge into a new node. However, this method is inefficient when applied to large graphs. Instead, NodePool only preserves high-scoring nodes and drops others in the next layer, which we focus on in this paper. The classical NodePool method named as TopkPool\cite{gao2019graph} learns the node score by projecting node features to a learnable vector, ignoring the graph structure and neighborhood information, leading to poor performance. Later on, some methods are proposed using multi-order information outperforms TopkPool, such as SAGPool\cite{lee2019self} and ASAP\cite{ranjan2020asap} and iPool\cite{gao2019ipool}, in which node scores are learned in consideration of the graph structure.


\paragraph{$k$-order convolution kernel} We call graph convolution using wider receptive fields (greater than 1) as $k$-order methods. The convolution between signal $X$ and the $k$-order filter $g_k$ on graph $\mathcal{G}$ is $X*_{\mathcal{G}}g_k=T_kX$. Here we introduce two types of $k$-order convolution kernel, which will be used latter. The first one is Chebyconv kernels, defined as

\begin{equation}
\label{eq:chebconv}
\begin{cases}
    & T_k = 2L T_{k-1} - T_{k-2}(L); \\
    & T_1 = L; \\
    & T_0 = I,\\
\end{cases}
\end{equation}

where $L$ is the normalized graph Laplacian\cite{defferrard2016convolutional}. Another efficient $k$-order convolution kernel is presented in Mixhop\cite{abu2019mixhop}:

\begin{equation}
\label{eq:mixhop}
    T_k=(D^{\frac{1}{2}}(A+I)D^{\frac{1}{2}})^k,
\end{equation}

where $A,D,I$ are adjacency matrix, degree matrix and identity matrix, respectively. It is noted that the self-loop terms $I$ is added in the normalized graph Laplacian, allowing  the $k$-order graph convolution to contain $(k-1)$-order message passing process.

\section{Methods}
Now we introduce the efficient yet expressive graph convolution and pooling for graph classification based on $k$-hops neighborhood information. In section 3.1, we illustrate the model architecture and common symbols. In section 3.2, we introduce the measure of $k$-hops information and provide the guidance of selecting $k$. In section 3.3, the light version of $k$-order graph convolution and pooling are separately described. In this paper, '$k$-order' equals to '$k$-hop', where the former describes convolution\&pooling and the latter is used for neighborhood information on graph.

\subsection{Model Architecture}
\begin{figure}[H]
    \centering
        \includegraphics[width=5 in]{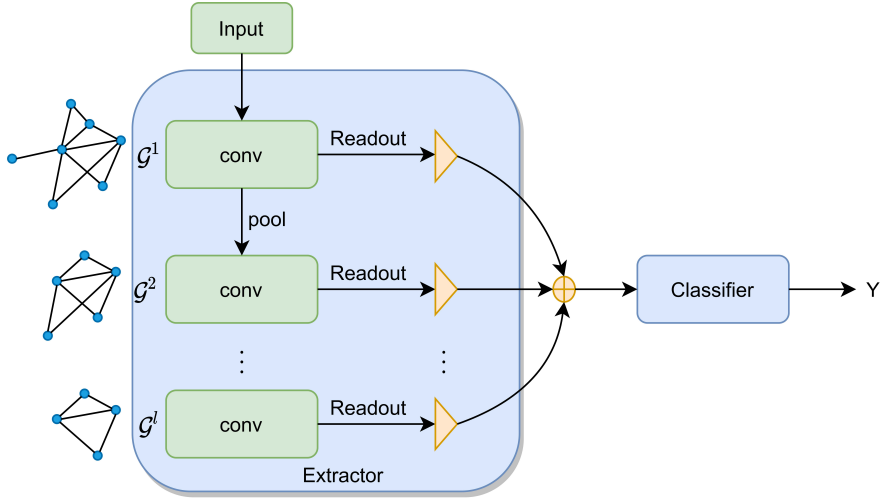}
	    \caption{Overall framework. The feature extractor is used to learn hierarchical graph features via stacked $k$-order convolution and pooling. Different levels graph signals are projected to a the same shape through the \textit{Readout} function and then added together. Finally, a classifier such as MLP generate the predicted label.}
	    \label{fig:overall_architecture}
\end{figure}

Denote the graph signal as $\mathcal{G}(V,E,X)$, where $V=\{v_i|i=1,2,\ldots,n\},E,X \in \mathbb{R}^{n,d},n,d$ are vertexes set, edge set, node features, number of nodes and the feature channels, with a corresponding adjacency matrix $A \in \mathbb{R}^{n,n}$.  As shown in Fig.\ref{fig:overall_architecture}, the graph classifier containing feature extractor and classifier mapping $\mathcal{G}(V,E,X)$ to $Y \in \{0,1,\dots,C\}$, where $C$ is the number of classes. The extractor learns hierarchical graph features through light $k$-order convolution and pooling along with readout function, and the classifier generate the predicted label using extracted features. Symbolically, the $k$-order convolution and pooling in layer $l$ is $Pool^{(l)}\circ Conv^{(l)}:\mathcal{G}^l(V^l,E^l,X^l) \mapsto \mathcal{G}^{l+1}(V^{l+1},E^{l+1},X^{l+1})$. For simplicity, we ignore the layer index $l$ and re-write $\mathcal{G}^l(V^l,E^l,X^l)$, $\mathcal{G}^{l+1}(V^{l+1},E^{l+1},X^{l+1})$ as $\mathcal{G}(V,E,X),\mathcal{G}^{'}(V^{'},E^{'},X^{'})$ in later sections. The function $Readout:\mathcal{G}(V,E,X) \mapsto \mathbb{R}^{d}$ indicates the global graph pooling, such as max pooling and mean pooling. The $Classifier:\mathbb{R}^{n,d'}\mapsto \{0,1,\dots,C\}$ can be chosen as MLP, where $d'$ is the output feature channels.

\subsection{Multi-hops Information}
When $k$-order conv\&pool is applied on graph $\mathcal{G}(V,E,X)$, a dilemma occurs, that is a small $k$ imposes restrictions on models' learning ability to aggregate message from farther nodes, while a large $k$ requires large number of parameters and high cost of computation. To balance the performance and efficiency, we propose a criterion to measure the information gains as $k$ increases and the strategy for selecting a smaller yet informative $k$. 

\paragraph{kInfo} For the graph signal $X \in \mathbb{R}^{n,d}$, the global and local feature density of the $c$-th channel are

\begin{align}
\centering
    \begin{cases}
        & G_{c}(x)=\hat{G}(x;X_{:,c});\\ %
        & L_{i,c}^{(k)}(x)=\frac{G_{c}(x)}{\sum_{x \in \mathcal{F}_{i,c}^k}{G_{c}(x)}}.\\
    \end{cases}
\end{align}

where $G_{c}$ is the global density over the $c$-th dimension (channel) of the input space, $\hat{G}$ is a density function estimated by $X_{:,c}$, and  $\mathcal{F}_{i,c}^k$ is the set of the $c$-th dimension of node features in $k$-hops graph centered on $i$, in which we define the local normalized density $L_{i,c}^{(k)}$. Gaussian KDE is used as the density estimation function in this paper. The entropy of $L_{i,c}^{(k)}$ is used to estimate the $k$-hops neighborhood information, so called kInfo of the $c$-th channel:

\begin{align}
\centering
  H_{i,c}^{(k)} = -\sum_{x \in \mathcal{F}_{i,c}^k}{ L_{i,c}^{(k)}(x) \log L_{i,c}^{(k)}(x) }.
\end{align}

\paragraph{Information Gains on k-hop graph} The probability density of kInfo estimated by $\{H_{i,c}^{(k)}| 1\leq i\leq n\}$ via the estimating function $\hat{P}$ is:
\begin{align}
\centering
  P_c^{(k)}(H)=\hat{P}(H; \{H_{i,c}^{(k)}| 1\leq i\leq n\}),
\end{align}
The \textbf{Information Gains}($IG(k)$) between the $(k-1)$-hop and $k$-hop can be measured by the average KL-divergence:

\begin{equation}
\label{eq:ID}
    IG(k)=\frac{1}{|d|} \sum_{c=1}^{d}{KL(P_{c}^{(k)}||P_{c}^{(k-1)})},k>0
\end{equation}


For completeness, we set $IG(0)=0$. Intuitively, larger $IG(k)$ indicates more difference between $P_c^{(k)}$ and $P_c^{(k-1)}$, and there are more information brought from the $(k-1)$-hop.

\paragraph{Selecting $k$} The total information gain from $k=1$ to $k=\hat{k}$ is $\sum_{k=1}^{\hat{k}}{IG(k)}$, where the information loss is $\sum_{k=0}^{\infty}{IG(k)}-\sum_{k=1}^{\hat{k}}{IG(k)}=\sum_{k=\hat{k}+1}^{\infty}{IG(k)}$. By introducing a small positive $\epsilon$ as the percent of information loss and assuming that $\sum_{k=0}^{\infty}{IG(k)}$ converges, $\hat{k}$ can be chosen as:

\begin{align}
\centering
    &\min \hat k; \quad \quad \quad  \text{s.t.} \quad  \frac{\sum_{k=\hat{k}+1}^{\infty}{IG(k)}}{\sum_{k=0}^{\infty}{IG(k)}} \leq \epsilon.
  \label{eq:selecting_k}
\end{align}


\subsection{Light Multi-order Convolution\&Pooling}

\begin{figure}[h]
    \centering
        \includegraphics[width=6in]{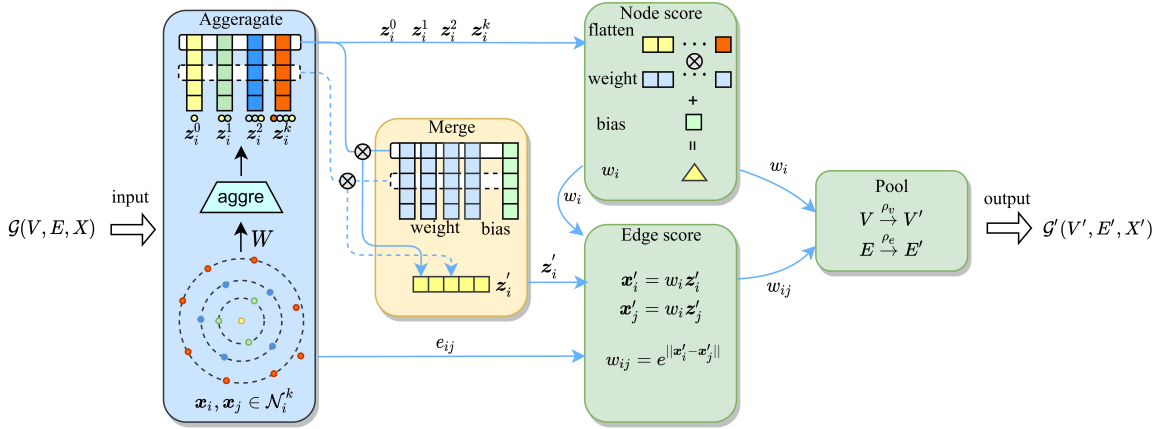}
	    \caption{The proposed convolution and pooling layer. In the lower left corner, $\boldsymbol{x}_i$ colored with yellow is the center node, $\mathcal{N}_i^k$ means the $k$-hops neighborhood of $v_i$ and neighbors in different hops are colored in different colors. The input feature vectors, such as $\boldsymbol{x}_i,\boldsymbol{x}_j \in \mathbb{R}^d$, are firstly projected to $\mathbb{R}^{d'}$. Then the projected features of $\{0,1,\dots,k\}$ hops are aggregated as anchor vector $\boldsymbol{z}_i^k$, who can express the $k$-hops information. By using the convolution operation of decoupling sum for different channels of anchor vectors, we can get the convolution output $\boldsymbol{z}'_i$. The importance score of node $v_i$ and edge $e_{ij}$ are $w_i$ and $w_{ij}$, where $w_i$ is learned from anchor vectors via attention mechanism and $w_{ij}$ is calculated from the difference of the two endpoints of $e_{ij}$. Finally, the pooling layer output the feature vectors $\boldsymbol{x}'_i=\boldsymbol{z}'_i/\text{norm}(\boldsymbol{z}'_i)\times w_i$ and only preserving nodes and edges with higher scores.}
	    \label{fig:ConvPool}
\end{figure}


\paragraph{$k$-order conv} The mapping path of graph convolution from $X$ to $X'$ is $\mathbb{R}^{n,d}\stackrel{Aggregate}{\longmapsto}\mathbb{R}^{n,k+1,d'} \stackrel{Merge}{\longmapsto}\mathbb{R}^{n,d'}$. The $Aggeragate:X\in \mathbb{R}^{n,d}\mapsto Z \in \mathbb{R}^{n,k+1,d'}$ uses convolution kernels mentioned in Eq.\ref{eq:chebconv} or Eq.\ref{eq:mixhop} to aggregate $k$-hops graph features as $Z=[Z^0||Z^1||\dots||Z^k],Z^k=T_k X W$, where $T_k\in \mathbb{R}^{n,n}$, $W\in \mathbb{R}^{d,d'}$ are convolution kernel and learnable feature transformation matrix, and $||$ indicates concatenation.The $i$-th row in $Z^k\in \mathbb{R}^{n,d'}$ is called a $k$-hop anchor vector $\boldsymbol{z}^k_i$ of $i$. For each vertex $i$, its anchor vectors $\boldsymbol{z}^0_i,\dots,\boldsymbol{z}^k_i \in \mathbb{R}^{d'}$ can be merged together by applying separate fully-connected in each channel:

\begin{align}
\centering
    \label{eq:graph_conv}
    \boldsymbol{z}'_i=\sum_k{\boldsymbol{z}_i^k \odot \boldsymbol{w}^k +\boldsymbol{b}},
\end{align}

where $\boldsymbol{w}^k, \boldsymbol{b} \in \mathbb{R}^{d'}$ are the learnable weight and bias and $\odot$ is element-wise product. Compared to $Chebyconv:\mathbb{R}^{n,(k+1)d} \mapsto \mathbb{R}^{n,d'}$, the $Merge\circ Aggregate: \mathbb{R}^{n,d}\mapsto\mathbb{R}^{n,k+1,d'}\mapsto\mathbb{R}^{n,d'}$ decouples the dimension of $k$ and $d$, and thus reduce the total trainable parameters for each convolution layer from $(k+1)\times d\times d'$ to $(d+k+1)\times d'$. The $Merge$ operation can be applied to Chebconv\cite{defferrard2016convolutional} in Eq.\ref{eq:chebconv} or MixhopConv\cite{abu2019mixhop} in Eq.\ref{eq:mixhop}, leading to their light version called LiCheb or LiMixhop.

\paragraph{$k$-order pool} Under the node scoring pooling framework, nodes with higher scores will be retained, otherwise they will be dropped. Previous works have shown that the wider neighborhood information is beneficial to learn node \cite{lee2019self,ranjan2020asap}, while their computation complexities are greatly increased. Here we use the $k$ anchor vectors to learn node scores more efficiently:

\begin{align}
\centering
    \label{eq:graph_conv}
    \begin{cases}
        & w_i=Relu(sum(\boldsymbol{\theta}\odot[\boldsymbol{z}_i^0||\boldsymbol{z}_i^1||\dots||\boldsymbol{z}_i^k])));\\
        & \boldsymbol{x}'_i=w_i \boldsymbol{z}'_i/(\boldsymbol{z}'_i),\\
    \end{cases}
\end{align}

where $w_i$ is the node score of vertex $i$, $\boldsymbol{x}'_i$ is the output feature of the pooling layer and $\boldsymbol{\theta} \in \mathbb{R}^{k\cdot d'}$ is attention weight for concatenated anchor vectors $[\boldsymbol{z}_i^0||\boldsymbol{z}_i^1||\dots||\boldsymbol{z}_i^k] \in \mathbb{R}^{k\cdot d'}$. Empirically, we find that the normalization of $\boldsymbol{z}'_i$ can improve the performance. For structure learning, we also consider the edge score $w_{ij}$ as feature difference of endpoints of $e_{i,j}$, i.e. $\boldsymbol{x}_i$ and $\boldsymbol{x}_j$. The more different the endpoints are, the more information they can get from each other, and the more important the edge is.

\begin{align}
\centering
    \label{eq:graph_conv}
    w_{ij}=e^{||\boldsymbol{x}_i-\boldsymbol{x}_j||}.
\end{align}

The nodes and edges with higher score, as well as the resulting subgraphs will be preserved on the next layer, written as

\begin{align}
\centering
    \label{eq:graph_pool}
    \begin{cases}
        & V' = Top_{\rho_{v}} (V) = \{ v_i| w_i\geq w(\rho_{v}), v_i\in V\};\\
        & E' = Top_{\rho_{e}} (E) = \{ e_{ij}| w_{ij}\geq w(\rho_{e}), v_i, v_j\in V', e_{i,j}\in E\},\\
    \end{cases}
\end{align}

where $\rho_{v},\rho_{e} \in (0,1]$ are the ratio for nodes and edges, $w(\rho_{v})$ and $w(\rho_{e})$ are the corresponding threshold values. For example, if $\rho_{v}=0.6$, $60\%$ nodes will be dropped.

\section{Experiments}
In section 4.1, we study the neighborhood information gain of 6 datasets and provide the guidance of selecting $k$. In section 4.2 and 4.3, we demonstrate the superiority of the proposed $k$-order convolution and pooling in terms of performance and complexity.

\paragraph{Basic setting} As shown Table.\ref{Table:dataset}, six benchmark datatsets for graph classification including PROTEINS, D\&D\cite{dobson2003distinguishing}, NCI1, NCI109, Multagenicity and FRANKENSTEIN are considered. Our code is based on PyTorch 1.7, runing on the Tesla V100(32G). Because some results of previous works are highly rely on the choice of random seeds and their experiment protocols are different, we re-test all the baseline modules under the same code architecture to ensure the fairness. We separate each dataset 10 times into training set($80\%$), validation set($10\%$) and testing set($10\%$), using ten different random seeds, and the average accuracy is reported in this paper. We train each model at most 500 epochs, and the early stop patient is 30 epochs.

\begin{table}[h]
\centering
\caption{The description of real-world datasets for evaluation. The columns indicate graph number($|\mathcal{G}|$), vertex number($|\mathcal{V}|$), edge number($|\mathcal{E}|$), average vertex(Avg.$|\mathcal{V}|$) or average edge(Avg.$|\mathcal{E}|$) number per graph, feature channels and class number($|C|$). }
    \begin{tabular}{cccccccc}
    \toprule
     Datasets   & $|\mathcal{G}|$    & $|\mathcal{V}|$   & $|\mathcal{E}|$     & Avg.$|\mathcal{V}|$    & Avg.$|\mathcal{E}|$  & features  & $|C|$ \\
    \midrule
    
    PROTEINS        & 1,113     & 43,471    & 81,044      & 39.0575     & 1.8643     & 4      & 2\\
    DD              & 1,178     & 334,925   & 843,046      & 284.3166    & 2.5171    & 89      & 2\\
    NCI1            & 4,110     & 122,747   & 132,753      & 29.8655     & 1.0815     & 37      & 2\\
    NCI109          & 4,127     & 122,494   & 132,604      & 29.6811     & 1.0825     & 38      & 2\\
    Multagenicity   & 4,337     & 131,488   & 133,447      & 30.3177     & 1.0149     & 14      & 2\\
    FRANKENSTEIN    & 4,337     & 73,283    & 77,534      & 16.8972     & 1.0580     &780      & 2\\
    \bottomrule
    \end{tabular}
\label{Table:dataset}
\end{table}

\subsection{Convolution Order Selection}
\paragraph{Experiment setting} The experiment aims to study the information gains (IG) between the $k$ and $(k-1)$ order and discuss the strategy of selecting $k$. Firstly, We use all the samples to calculate $\{IG(k)|2\leq k \leq 10\}$ for each dataset according to Eq.\ref{eq:ID}. Curve fitting is performed to get the approximated mathematical expression of $IG(k)$. Besides, we further analyze the property of $IG(k)$ and propose the guidance of selecting $k$.

\begin{figure}[h]
\parbox{0.52\textwidth}{
    \centering
    \includegraphics[width=0.9\linewidth]{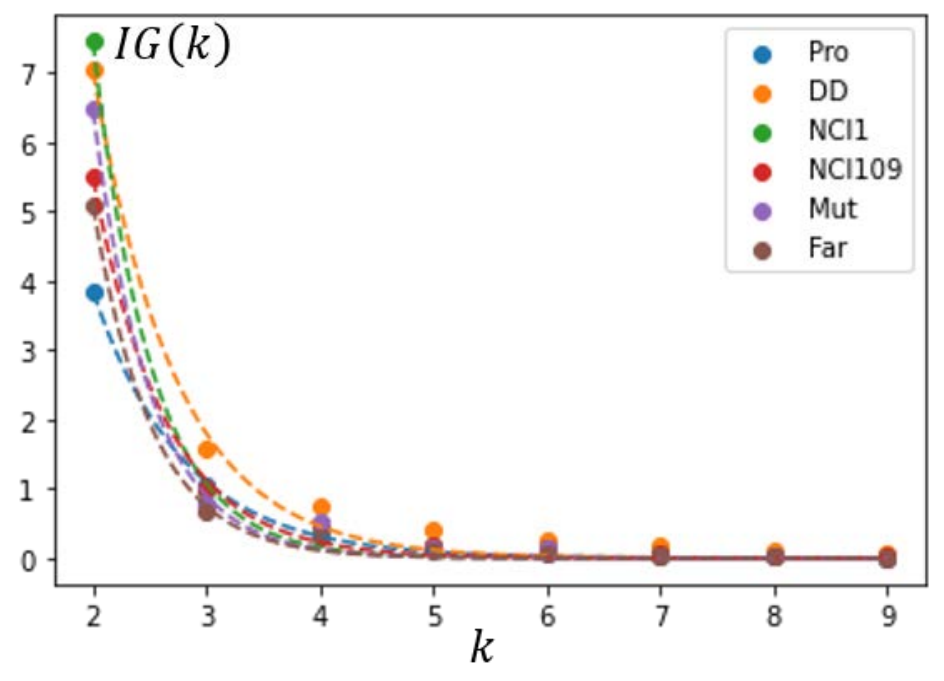}
    \caption{$IG(k)$ over different datasets, where the first axis indicates the number $k$ and the second axis is $IG(k)$. $IG(k)$ is consistently to be a decreasing function over different datasets, converging to 0 with exponential speed. }
    \label{fig:IG_k}
}
\parbox{0.5\textwidth}{
    \centering
    \begin{tabular}{ccccc}
        \toprule
         Datasets   & a     &b     &  R$^2$ score  & MSE\\
        \midrule
        PRO     & 46.8505  & 1.2501 & 0.9995 & 0.0007\\
        DD      & 107.0625 & 1.3622 & 0.9915 & 0.0422\\
        NCI1    & 356.5393 & 1.9339 & 0.9965 & 0.0200\\
        NCI109  & 130.8210 & 1.5853 & 0.9952 & 0.0149\\
        Mul     & 335.1961 & 1.9743 & 0.9943 & 0.0246\\
        FRA     & 235.0988 & 1.9169 & 0.9976 & 0.0066\\
        \bottomrule
    \end{tabular}
    \caption{Curve fitting results of $IG(k)=ae^{-bk}$, which are pretty good in terms of $R^2$ score and MSE. Because $b>1$, $\sum_{k=0}^{\infty}{IG(k)}$ converges, thus the assumption on Eq.\ref{eq:selecting_k} is valid.}
    \label{Table:IG_k}
}
\end{figure}

\begin{figure}[htbp]
    \centering
    \subfigure
        {
        	\begin{minipage}{8cm}
        	\centering          
        	\includegraphics[width=3.5 in]{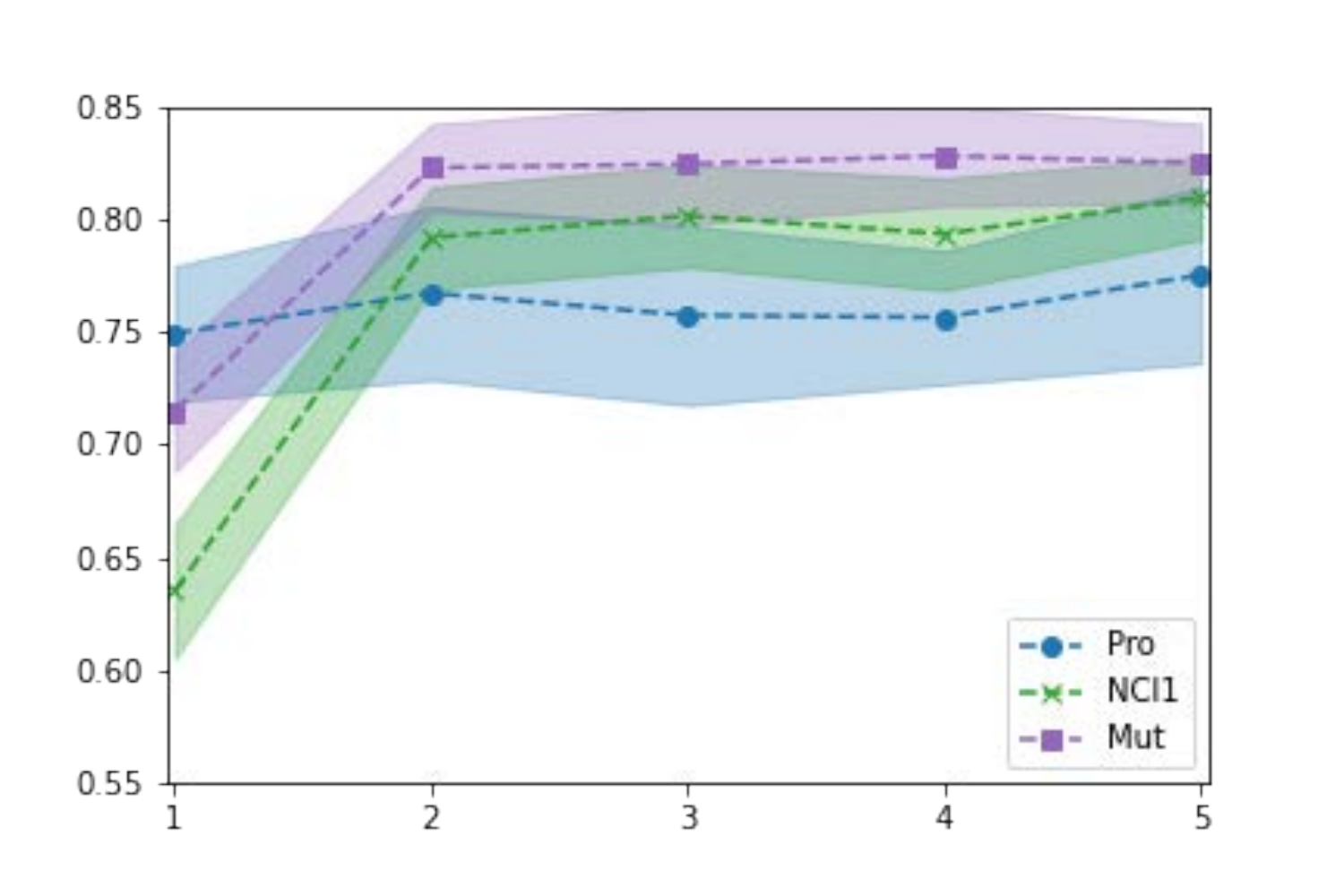}   
        	\end{minipage}
        }
    \subfigure
        {
        	\begin{minipage}{8cm}
        	\centering          
        	\includegraphics[width=3.5 in]{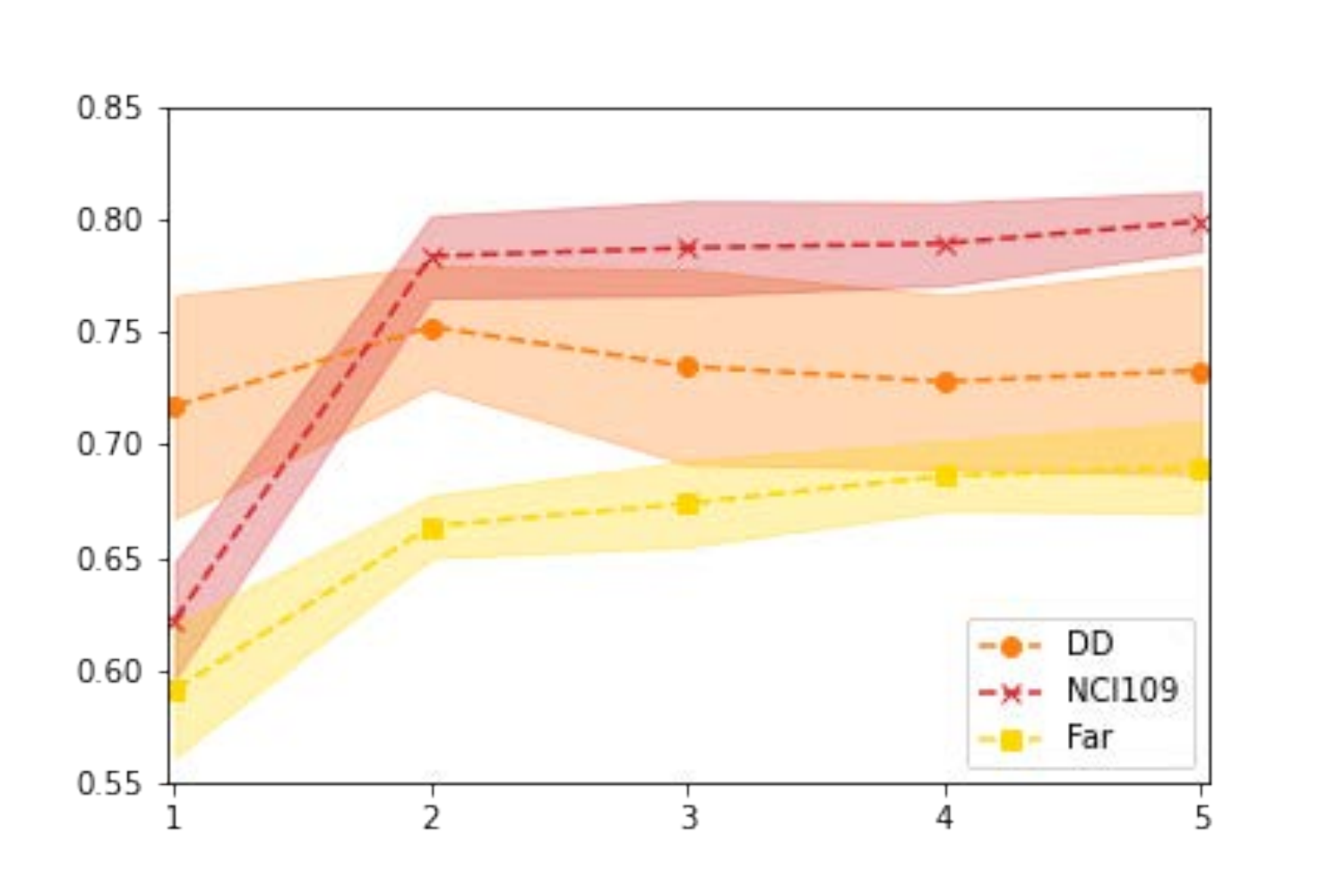}   
        	\end{minipage}
        }
    \caption{The average accuracy of LiCheb convolution(without pooling) under different $k$ across 10 random seeds. Limited to the computation cost, $k$ is no more than 5. As we can see $k$-order($k>1$) convolution can always surpass $1$-order, while the performance gain vanished fast, similar to the information gain. }
    \label{fig:acc_k}
\end{figure}


    

\paragraph{Results and discussion} In Fig.\ref{fig:IG_k} and Table.\ref{Table:IG_k}, we present $IG(k)$ and its curve fitting results. The average accuracy of LiCheb convolution across different $k$ is shown in Fig.\ref{fig:acc_k}. 

\begin{itemize}
    \item The fitting results of $IG(k)=ae^{-bk},a,b>0$ are pretty good in terms of R$^2$ score and MSE, seeing Table.\ref{Table:IG_k}, demonstrating empirically that the information difference converges to 0 with a negative \dotuline{exponential speed} with respect to $k$. 
    \item  It indicates $k$ can be limited to a smaller range, avoiding the expensive computation complexity of larger $k$ for practical scenarios. With $IG(k)=ae^{-bk}$, the information loss percent defined in Eq.\ref{eq:selecting_k} can be simplified as $ \epsilon = \frac{\sum_{k=\hat{k}+1}^{\infty}ae^{-bk}}{\sum_{k=1}^{\infty}ae^{-bk}} = e^{-b\hat{k}}$, and thus choosing $\hat k$ for convolution indicates that the information loss equals to $e^{-b\hat k} \times 100\%$.  For example, when choosing the suitable $k$ for multi-order convolution on PRO dataset, $\hat{k}=3$ is corresponding to only $e^{-3 \times 1.25}=0.0235 = 2.35\%$ information loss percent, and thus $k$ can be selected within 3 in our experiments. 

    \item The average accuracy in Fig.\ref{fig:acc_k} is highly related to kInfo, as the accuracy improves dramatically when a small $\hat k$ increases, while performance will fluctuate steadily when $\hat k$ reaches to certain value. 
\end{itemize}

\subsection{Multi-order Convolution}
\paragraph{Experiment setting}
We evaluate LiCheb and LiMixhop on 6 public benchmarks of graph classification together with baseline models under the same condition,i.e., code\&network architecture, random seeds, batch size, optimizer. For ensuring the reproducibility, important hyper-parameters can be found in Table.\ref{Table:search_space}. The convolution methods include GCN\cite{kipf2016semi}, GAT\cite{velivckovic2017graph}, Chebconv\cite{defferrard2016convolutional} and Mixhop\cite{abu2019mixhop}, the first two of which are 1-order methods, and the last two are $k$-order methods.

\begin{table}[h]
\centering
\caption{Hyper parameters of six datasets. Use ';' to separate different values of the parameter on various datasets.}
    \begin{tabular}{cccccccccc}
    \toprule
    Param name   & $k$   & network layer  &batch size & lr   & $\rho_{v}$(node pooling)    &  $\rho_{e}$(edge pooing)  \\
    \midrule
    Search space   & 2    & 5  & 256 & 0.001    & $\{0.6;0.8;0.9;0.9;0.9;0.9\}$     & $\{0.8;0.7;0.9;0.4;0.7;0.6\}$ \\
    \bottomrule
    \end{tabular}
\label{Table:search_space}
\end{table}

\setlength{\tabcolsep}{0.9mm}{
\begin{table}[h]
    \centering
    \caption{Results of $k$-order convolution. The number of model parameters and the run time over 50 epoch are counted in experiments of NCI1. }
    \begin{tabular}{cccccccccc}
    \toprule
    Categories                & Pro     & D\&D         & NCI1         & NCI109       & Mut & Far  & params & time\\
    \midrule
    GCN       & $74.8 \pm 2.5$ & $73.9 \pm 4.1$ & $75.8 \pm 2.3$ & $74.5 \pm 1.8$ & $79.6 \pm 3.2$   & $63.8 \pm 2.5$ & 112.2k & 58.5s\\
    GAT       & $74.3 \pm 3.5$ & $ 73.6 \pm 5.1 $ & $76.4 \pm 4.4$ & $75.6 \pm 1.7$ & $79.9 \pm 1.6$   & $59.9 \pm 1.7$   & 113.5k & 79.3s\\
    \midrule
    Mixhop & $73.5 \pm 3.2$ & $ 75.9 \pm 4.9 $ & $76.0 \pm 4.9$ & $74.5 \pm 2.8$ & $81.4 \pm 2.3$   & $62.5 \pm 1.9$   & 120.5k & 66.1s\\
    LiMixhop & $\bm{73.9 \pm 3.7}$ & $ \bm{76.1 \pm 6.0} $ & $\bm{78.4 \pm 2.2}$ & $\bm{77.3 \pm 2.3}$ & $\bm{82.3 \pm 2.0}$   & $\bm{63.9 \pm 1.8}$   & 120.5k & 68.9s\\
    \midrule
    Chebconv & $73.7 \pm 3.6$ & $\bm{77.3 \pm 3.7}$ & $78.3 \pm 3.0$ & \bm{$78.1 \pm 2.5$} & $80.8 \pm 2.7$   &  \bm{$66.1 \pm 2.2$}  &254.4k & 73.2s\\
    LiCheb   & $\bm{74.6 \pm 4.5}$ & $76.2 \pm 4.9$ & \bm{$78.6 \pm 1.6$} & $76.6 \pm 3.0$ & \bm{$81.4 \pm 1.8$}  &   $65.9 \pm 2.2$    &120.5k & 74.6s \\
    \bottomrule
    \end{tabular}
    \label{Table:conv_results}
\end{table}
}

\paragraph{Results and discussion} 
The average accuracy over 10 different random seeds along with the model parameters and runtime are reported in Table.\ref{Table:conv_results}. The first group shows results of $1$-order convolution, and the second and third group show the results of $k$-order convolution compared with the proposed method. 
\begin{itemize}
    \item By comparing groups, we conclude that $k$-order graph convolution strongly outperform 1-order methods except for the Protein dataset.
    \item Through group 2, we can see LiMixhop achieves higher performance under the similar parameters, compared to origin Mixhop.
    \item From group 3, we observe that LiCheb can reduce the parameter more than half under the similar performance, thus eabling more efficient computation.
\end{itemize}

\paragraph{Expressive ability} Here we state that 1 layer $k$-order convolution is more expressive than $k$ layer stacked 1-order convolutions. According to the result of Wu\cite{wu2019simplifying}, $k$ layer GCN can be simplified as $\bm{Y}={softmax}(A^k X \Theta)$, without negatively impact on accuracy. Thus $k$-hops graph signal $F=\sum_k{\alpha_k A^k X},\alpha_k \neq 0$ cannot totally be represented by GCN. When approximating the target signal, the lower order information $\sum_{k-1}{\alpha_k A^k X}$ cannot be ignored in approximation theory.

\subsection{Multi-order Pooling}
\paragraph{Experiment setting} 
We compare the proposed pooling layer with baseline modules in terms of performance and efficiency. The latest pooling baselines such as TopkPool\cite{gao2019graph}, SAGPool\cite{lee2019self}, ASAP\cite{ranjan2020asap} and EdgePool\cite{diehl2019edge} are chosen as baseline modules, where the first three are the NodePool methods and the fourth is EdgePool method. We further do ablation study for the normalized feature(NF), node pooling(pN) and edge pooling(pE) to show whether these operation can obtain the performance gain. For fair comparison, all the experiments use LiCheb as the convolution module, and the only difference among different experiments is the pooling method. Other experiment\&data setting keep the same as Sec.4.2. Hyper-parameters can still be found in Table.\ref{Table:search_space}.

\setlength{\tabcolsep}{0.9mm}{
\begin{table}[h]
    \centering
    \caption{Pooling results. The results of APAP on D\&D is missing because it requires too many GPU memory, even we have reduced the batch size. The parameter numbers and run time over 50 epoch are counted in experiments of NCI1.}
    \begin{tabular}{ccccccccc}
    \toprule
    Categories                & Pro     & D\&D         & NCI1         & NCI109       & Mut & Far  & params & time\\
    \midrule
    TopkPool    & $73.1\pm3.1$  & $68.1\pm4.7$  & $76.1\pm2.3$  & $75.3\pm2.4$  & $79.6\pm2.7$    &$61.5\pm2.8$  & 120.6k & 264.4s\\
    SAGPool     & $74.6\pm7.1$  & $74.8\pm4.0$  & $75.0\pm1.5$  & $76.0\pm2.3$  & $78.1\pm1.9$    & $64.0\pm2.8$ & 120.8k & 264.9s\\
    EdgePool    &$74.7\pm5.2$   & $76.4\pm4.8$  & $76.4\pm2.5$  & $\bm{76.8\pm1.8}$  & $80.3\pm2.5$    & $64.6\pm3.0$ & 120.8k & 387.9s\\
    ASAP        & $75.1\pm3.8$  & $--$          & $76.9\pm1.8$  & $75.9\pm1.9$  & $\bm{80.7\pm1.9}$    & $64.2\pm2.8$ & 137.7k & 398.9s\\
    \midrule
    Our(noNF+pN) & $72.8\pm5.0$ & $76.7\pm4.6$  & $76.3\pm1.8$  & $75.4\pm3.0$    & $78.8\pm 2.3$& $63.1\pm2.1$ & 116.7k & 166.3s\\
    Our(NF+pN) &$76.3\pm4.6$   & $76.2\pm4.9$   & $77.4\pm2.4$  & $76.1\pm1.9$  & $78.7\pm1.7$    & $64.5\pm1.2$ & 116.7k & 201.0s\\
    Our(NF+pN+pE)    &$\bm{76.3\pm4.5}$   & $\bm{76.4\pm4.1}$   & $\bm{78.2\pm2.5}$  & $76.5\pm2.0$  & $80.3\pm2.8$    & $\bm{65.0\pm1.9}$ & 116.7k & 196.3s\\
    \bottomrule
    \end{tabular}
    \label{Table:pool_results}
\end{table}
}

\paragraph{Results and discussion} 
In Table.\ref{Table:pool_results}, we report the average accuracy of different pooling methods under 10 different random seeds, with baseline results shown in group 1 and the results of the proposed method and ablation study shown in group 2.

\begin{itemize}
    \item  Comparing group 1 and the proposed(NF+pN+pE), conclusion are reached that the proposed pooling can achieve the state-of-the-art performance on most of these datasets while smaller number of parameters with less computational cost are used.
    \item From (noNF+pN) and (NF+pN) in group 2, we can see that the normalized feature (NF), who projecting the output of convolution to the hyper-sphere, can increase the accuracy on these datasets except for D\&D and Multagenicity. Because it replaces vector norm as the node importance score, pooling methods based on the node score can work better.
    \item From (NF+pN) and (NF+pN+pE), it can be proved that the structure learning(edge dropping) can further improve the performance, which indicates there are redundant edges in the original dataset, and edges with large endpoint difference can bring more information gain. All in all, the proposed pooling method can achieve SOTA performance in a more efficient way.
\end{itemize}

\paragraph{Rethink pooling} 
Under the strict testing protocol, we also found that GNNs with graph pooling could not exceed GNNs using convolution only in all these benchmarks, which is consistent to Mesquita\cite{mesquita2020rethinking}. One of the reasons is that the graph structure is broken after pooling. However, additional complex structure learning module requires much more computation cost. Nevertheless, we believe that pooling is necessary for the concern of efficiency when the graph scales to thousands of nodes. When and how we shall use graph pooling? This is a question worth revisiting, left for future research.

\section{Conclusion}
In this paper, we investigate how to design efficient $k$-order convolution and pooling module for graph classification tasks. We first propose a criterion of $k$-hop neighborhood information(kInfo) to guide the selection of $k$. Then we propose efficient $k$-order convolution and pooling requiring few additional parameters while significantly improving the performance. Comprehensive and fair experiments on six widely used graph classification benchmarks show our superior compared to a other SOTA algorithms in terms of performance and efficiency.


\bibliographystyle{unsrt}  

\newpage
\newpage
\bibliography{references}

\begin{thebibliography}{10}

\bibitem{krizhevsky2017imagenet}
Alex Krizhevsky, Ilya Sutskever, and Geoffrey~E Hinton.
\newblock Imagenet classification with deep convolutional neural networks.
\newblock {\em Communications of the ACM}, 60(6):84--90, 2017.

\bibitem{defferrard2016convolutional}
Micha{\"e}l Defferrard, Xavier Bresson, and Pierre Vandergheynst.
\newblock Convolutional neural networks on graphs with fast localized spectral
  filtering.
\newblock {\em arXiv preprint arXiv:1606.09375}, 2016.

\bibitem{atwood2016diffusion}
James Atwood and Don Towsley.
\newblock Diffusion-convolutional neural networks.
\newblock In {\em Advances in neural information processing systems}, pages
  1993--2001, 2016.

\bibitem{kipf2016semi}
Thomas~N Kipf and Max Welling.
\newblock Semi-supervised classification with graph convolutional networks.
\newblock {\em arXiv preprint arXiv:1609.02907}, 2016.

\bibitem{velivckovic2017graph}
Petar Veli{\v{c}}kovi{\'c}, Guillem Cucurull, Arantxa Casanova, Adriana Romero,
  Pietro Lio, and Yoshua Bengio.
\newblock Graph attention networks.
\newblock {\em arXiv preprint arXiv:1710.10903}, 2017.

\bibitem{morris2019weisfeiler}
Christopher Morris, Martin Ritzert, Matthias Fey, William~L Hamilton, Jan~Eric
  Lenssen, Gaurav Rattan, and Martin Grohe.
\newblock Weisfeiler and leman go neural: Higher-order graph neural networks.
\newblock In {\em Proceedings of the AAAI Conference on Artificial
  Intelligence}, volume~33, pages 4602--4609, 2019.

\bibitem{gao2019graph}
Hongyang Gao and Shuiwang Ji.
\newblock Graph u-nets.
\newblock {\em arXiv preprint arXiv:1905.05178}, 2019.

\bibitem{zhang2019hierarchical}
Zhen Zhang, Jiajun Bu, Martin Ester, Jianfeng Zhang, Chengwei Yao, Zhi Yu, and
  Can Wang.
\newblock Hierarchical graph pooling with structure learning.
\newblock {\em arXiv preprint arXiv:1911.05954}, 2019.

\bibitem{lee2019self}
Junhyun Lee, Inyeop Lee, and Jaewoo Kang.
\newblock Self-attention graph pooling.
\newblock {\em arXiv preprint arXiv:1904.08082}, 2019.

\bibitem{gao2019ipool}
Xing Gao, Hongkai Xiong, and Pascal Frossard.
\newblock ipool--information-based pooling in hierarchical graph neural
  networks.
\newblock {\em arXiv preprint arXiv:1907.00832}, 2019.

\bibitem{ranjan2020asap}
Ekagra Ranjan, Soumya Sanyal, and Partha~P Talukdar.
\newblock Asap: Adaptive structure aware pooling for learning hierarchical
  graph representations.
\newblock In {\em AAAI}, pages 5470--5477, 2020.

\bibitem{diehl2019edge}
Frederik Diehl.
\newblock Edge contraction pooling for graph neural networks.
\newblock {\em arXiv preprint arXiv:1905.10990}, 2019.

\bibitem{bruna2013spectral}
Joan Bruna, Wojciech Zaremba, Arthur Szlam, and Yann LeCun.
\newblock Spectral networks and locally connected networks on graphs.
\newblock {\em arXiv preprint arXiv:1312.6203}, 2013.

\bibitem{niepert2016learning}
Mathias Niepert, Mohamed Ahmed, and Konstantin Kutzkov.
\newblock Learning convolutional neural networks for graphs.
\newblock In {\em International conference on machine learning}, pages
  2014--2023, 2016.

\bibitem{gilmer2017neural}
Justin Gilmer, Samuel~S Schoenholz, Patrick~F Riley, Oriol Vinyals, and
  George~E Dahl.
\newblock Neural message passing for quantum chemistry.
\newblock {\em arXiv preprint arXiv:1704.01212}, 2017.

\bibitem{hamilton2017inductive}
Will Hamilton, Zhitao Ying, and Jure Leskovec.
\newblock Inductive representation learning on large graphs.
\newblock In {\em Advances in neural information processing systems}, pages
  1024--1034, 2017.

\bibitem{ying2018hierarchical}
Zhitao Ying, Jiaxuan You, Christopher Morris, Xiang Ren, Will Hamilton, and
  Jure Leskovec.
\newblock Hierarchical graph representation learning with differentiable
  pooling.
\newblock In {\em Advances in neural information processing systems}, pages
  4800--4810, 2018.

\bibitem{abu2019mixhop}
Sami Abu-El-Haija, Bryan Perozzi, Amol Kapoor, Nazanin Alipourfard, Kristina
  Lerman, Hrayr Harutyunyan, Greg~Ver Steeg, and Aram Galstyan.
\newblock Mixhop: Higher-order graph convolutional architectures via sparsified
  neighborhood mixing.
\newblock {\em arXiv preprint arXiv:1905.00067}, 2019.

\bibitem{dobson2003distinguishing}
Paul~D Dobson and Andrew~J Doig.
\newblock Distinguishing enzyme structures from non-enzymes without alignments.
\newblock {\em Journal of molecular biology}, 330(4):771--783, 2003.

\bibitem{wu2019simplifying}
Felix Wu, Tianyi Zhang, Amauri Holanda~de Souza~Jr, Christopher Fifty, Tao Yu,
  and Kilian~Q Weinberger.
\newblock Simplifying graph convolutional networks.
\newblock {\em arXiv preprint arXiv:1902.07153}, 2019.

\bibitem{mesquita2020rethinking}
Diego Mesquita, Amauri Souza, and Samuel Kaski.
\newblock Rethinking pooling in graph neural networks.
\newblock {\em Advances in Neural Information Processing Systems}, 33, 2020.

\end{thebibliography}
\end{document}